\documentclass[times, review, 10pt]{elsarticle}

\usepackage{amssymb}
\usepackage{url}
\usepackage{amsmath}
\usepackage{tabularray}
\usepackage{booktabs}
\usepackage{array}
\usepackage{multirow}
\usepackage{makecell}
\usepackage{amsmath}
\usepackage{graphicx}
\usepackage{caption}
\usepackage[normalem]{ulem}
\usepackage{tabularray}
\usepackage{tabularray}
\usepackage{ulem}
\usepackage{multirow}
\usepackage{makecell} 
\usepackage{array}    
\usepackage{makecell}
\usepackage{hyperref}
\useunder{\uline}{\ul}{}
\usepackage{multirow}
\usepackage{makecell}
\usepackage{caption}

\renewcommand{\arraystretch}{1.2} 
\journal{Pattern Recognition}

\begin{document}

\begin{frontmatter}



\title{MELLM: A Flow-Guided Large Language Model for Micro-Expression Understanding}

\author[1,2]{Sirui Zhao\corref{cor1}\fnref{equal}}
\ead{siruit@ustc.edu.cn}
\author[1,2]{Zhengye Zhang\fnref{equal}}
\author[1,2]{Shifeng Liu}
\author[1,2]{Xinglong Mao}
\author[1,2]{\mbox{Shukang Yin}}
\author[3]{\mbox{Chaoyou Fu}}
\author[1,2]{Tong Xu\corref{cor1}}
\author[1,2]{Enhong Chen\corref{cor1}}

\fntext[equal]{These authors contributed equally to this work.}
\cortext[cor1]{Corresponding author}

\affiliation[1]{organization={University of Science and Technology of China},
            city={Hefei},
            postcode={230027}, 
            state={Anhui},
            country={China}}

\affiliation[2]{organization={State Key Laboratory of Cognitive Intelligence},
            city={Hefei},
            postcode={230027}, 
            state={Anhui},
            country={China}}

\affiliation[3]{organization={Nanjing University},
            city={Suzhou},
            postcode={215163},
            state={Jiangsu},
            country={China}}
            
\begin{abstract}
Micro-expressions (MEs), brief and low-intensity facial movements revealing concealed emotions, are crucial for affective computing. Despite notable progress in ME recognition, existing methods are largely confined to discrete emotion classification, lacking the capacity for comprehensive ME Understanding (MEU), particularly in interpreting subtle facial dynamics and underlying emotional cues. While Multimodal Large Language Models (MLLMs) offer potential for MEU with their advanced reasoning abilities, they still struggle to perceive such subtle facial affective behaviors. To bridge this gap, we propose a ME Large Language Model (MELLM) that integrates optical flow-based sensitivity to subtle facial motions with the powerful inference ability of LLMs. Specifically, an iterative, warping-based optical-flow estimator, named MEFlowNet, is introduced to precisely capture facial micro-movements. For its training and evaluation, we construct MEFlowDataset, a large-scale optical-flow dataset with 54,611 onset-apex image pairs spanning diverse identities and subtle facial motions. Subsequently, we design a Flow-Guided Micro-Expression Understanding paradigm. Under this framework, the optical flow signals extracted by MEFlowNet are leveraged to build MEU-Instruct, an instruction-tuning dataset for MEU. MELLM is then fine-tuned on MEU-Instruct, enabling it to translate subtle motion patterns into human-readable descriptions and generate corresponding emotional inferences. Experiments demonstrate that MEFlowNet significantly outperforms existing optical flow methods in facial and ME-flow estimation, while MELLM achieves state-of-the-art accuracy and generalization across multiple ME benchmarks. To the best of our knowledge, this work presents two key contributions: MEFlowNet as the first dedicated ME flow estimator, and MELLM as the first LLM tailored for MEU\footnote{Code is available at \url{https://github.com/zyzhangUstc/MELLM}}.
\end{abstract}







\end{frontmatter}



\section{Introduction}
Micro-expressions (MEs) are brief and involuntary facial expressions that occur when individuals attempt to conceal their genuine emotions, revealing underlying feelings despite conscious efforts to suppress them~\cite{ekman1969nonverbal}. The accurate recognition and understanding of MEs hold significant application value in fields such as forensic interrogation~\cite{yuan2025deception}, education\cite{sailer2021linking}, and healthcare~\cite{datz2019interpretation}. However, due to MEs' fleeting duration (typically less than 0.5 seconds~\cite{yan2013fast}), low intensity, and localized muscle movements, improving AI models' ability to perceive these fine-grained facial motions has been a major challenge in automatic ME recognition (MER). 

\begin{figure}[t]
    \centering
    \includegraphics [width=1\linewidth] {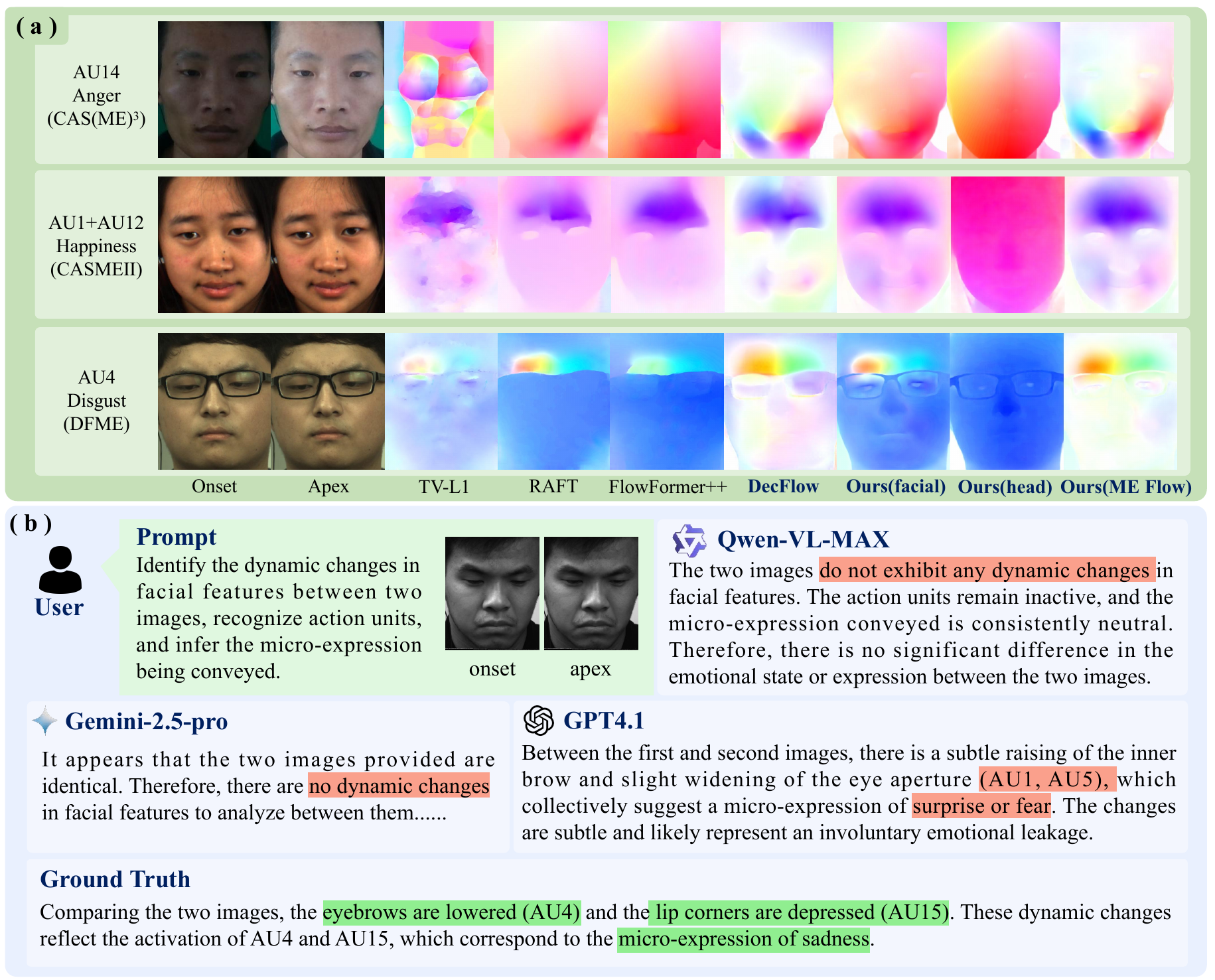}
    \caption{Limitations of optical-flow algorithms and MLLM for MEU. (a) Optical-flow estimates produced by various methods.  (b) Examples of several state-of-the-art MLLMs on MEU tasks.}
    \label{fig:figure1}
\end{figure}

A common paradigm in MER involves extracting optical flow representations from raw video sequences to capture fine-grained motion features for emotion classification~\cite{ZHOU2022108275, wang2024htnet}. Although these methods have demonstrated improved recognition accuracy, they still exhibit two major limitations. First, existing optical flow estimators often fail to adequately capture the subtle dynamics of MEs. As illustrated in \autoref{fig:figure1}(a), conventional methods such as TV-L1~\cite{werlberger2011optical} are sensitive to illumination variations~\cite{zhai2021optical}, primarily due to violations of the brightness constancy assumption. Although deep learning-based predictors~\cite{teed2020raft,shi2023flowformer++} have been developed to address this issue, they still struggle to reliably capture subtle facial movements and often remain vulnerable to global head motion. Even specialized models like DecFlow~\cite{lu2024facialflownet}, which explicitly attempt to compensate for head movement, lack robustness in practical settings, such as when subjects wear glasses, and provide limited accuracy in characterizing ME motion patterns. Second, while current MER systems achieve reasonable performance in emotion classification, they typically output only an emotion label without offering explicit or intuitive analysis of the nuanced facial movements and affective subtleties underlying MEs. This lack of detailed motion reasoning ultimately limits their interpretability and practical applicability.

Recent multimodal large language models (MLLMs) have demonstrated exceptional performance across a diverse range of multimodal understanding and generation tasks by integrating visual, textual, and, in some cases, auditory inputs~\cite{Shukang,fu2025vita}. This progress opens up new possibilities for comprehensive ME understanding (MEU), including the interpretation of subtle facial dynamics and their underlying emotional cues. A key strength of MLLMs is their ability to recast traditional discriminative tasks, such as regression and classification, as text-generation problems. Consequently, harnessing the robust visual perception, multimodal reasoning, and language generation capabilities of MLLMs could facilitate a more comprehensive and interpretable analysis of MEs.

However, existing MLLMs are not well-equipped to capture the delicate, frame-level facial dynamics of MEs due to their subtle, transient, and spatially localized nature \cite{lian2024gpt}. Our preliminary experiments reveal that state-of-the-art (SOTA) MLLMs often treat the onset and apex frames in ME sequences as nearly identical, or they produce erroneous and potentially misleading outputs. As illustrated in \autoref{fig:figure1}(b), prominent models like Qwen-VL-MAX \cite{bai2025qwen2} and Gemini-2.5-pro \cite{team2023gemini} fail to detect the dynamic changes between onset and apex frames, while GPT-4.1 \cite{openai_gpt4.1_2025} generates incorrect Action Unit (AU) labels and expression inferences. The primary reason for this issue is the vision encoders, which are typically pretrained on tasks that prioritize coarse-grained visual semantics (e.g., object recognition, scene understanding). As a result, they are not sensitive to the subtle, short-lived, and spatially localized motions that characterize MEs. Furthermore, this core weakness cannot be easily rectified by direct fine-tuning, as the scarcity of labeled ME data makes this approach ineffective. Such training tends to either underfit the fine-grained motion cues or overfit to idiosyncratic appearance patterns, failing to substantially improve the model's fundamental sensitivity to low-level motion.

\begin{figure}[t]
    \centering
    \includegraphics [width=1\linewidth] {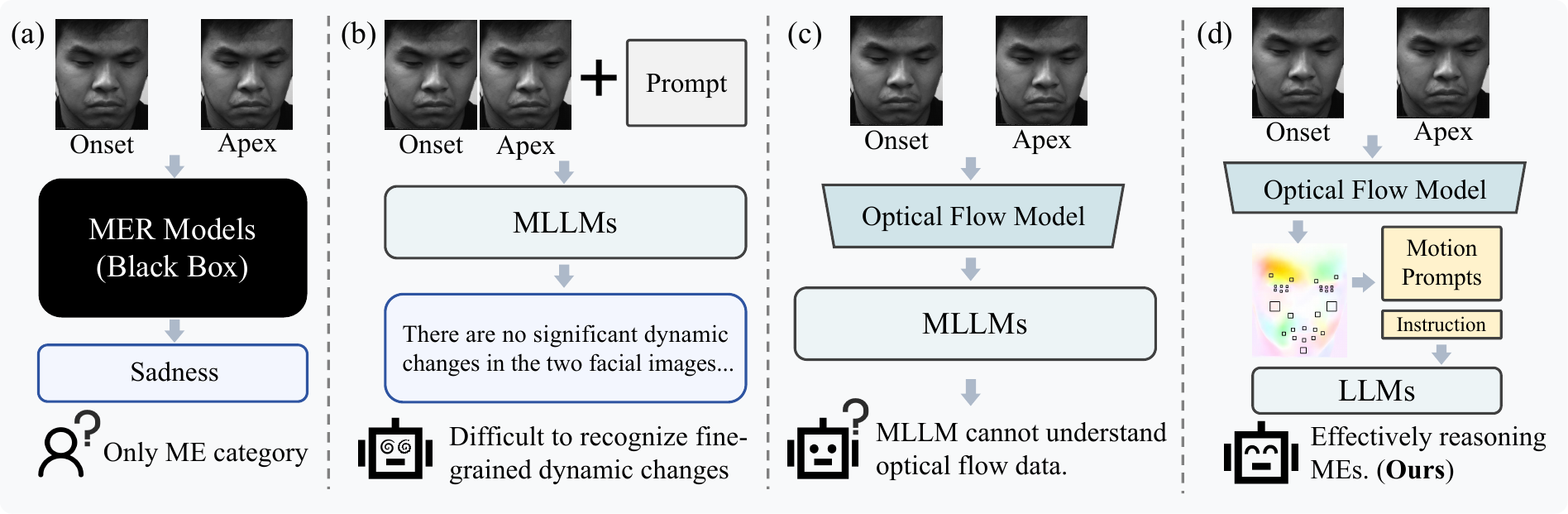}
    \caption{An overview of our proposed approach compared to other methods. (a) Traditional MER models act as black boxes, providing uninterpretable results. (b) MLLMs struggle to perceive the fine-grained dynamic changes inherent in MEs. (c) MLLMs are unable to directly process optical flow data or its visualizations. (d) Our approach converts optical flow from key ROIs into a structured Motion Prompt, enabling the LLM to effectively reason about the underlying ME.}
    \label{fig:figure2}
\end{figure}

To address these challenges, we propose the Micro-Expression Large Language Model (MELLM), a novel framework that integrates optical-flow-based motion sensitivity with the powerful inference capabilities of LLMs. \autoref{fig:figure2} illustrates a comparison between our proposed framework and other methods. Our approach encodes optical-flow signals from facial ROIs into structured motion prompts, enabling an LLM to natively understand and reason about these subtle movements.
Specifically, we first construct MEFlowDataset, a large-scale optical-flow dataset containing 54,611 onset-apex image pairs with diverse identities and subtle facial motions. Based on this dataset, we design MEFlowNet, an iterative optical-flow estimator that uses a warping-based approach to predict ME flow as the residual motion after compensating for head flow. As shown in \autoref{fig:figure1}(a), the ME flow estimated by MEFlowNet captures facial micro-movements more accurately than traditional or existing deep-learning methods.

We then introduce the Flow-Guided Micro-Expression Understanding (FGMU) paradigm. This paradigm encompasses two primary stages: first, the conversion of quantitative optical flow from MEFlowNet into structured motion prompts for each ROI; and second, a structured three-stage analysis of these prompts to infer the corresponding emotions. This strategy offers three main advantages: (1) it removes pixel-level redundancies and appearance-related confounders; (2) it provides the LLM with direct, interpretable motion cues for reasoning; and (3) it is robust to small dataset sizes, as the LLM can leverage its pretrained knowledge to map motion-related prompts to semantic inferences, obviating the need to relearn low-level visual signals.

Under this paradigm, we convert the ME Flow extracted by MEFlowNet into structured motion prompts, which are then used to construct MEU-Instruct, an instruction-tuning dataset for MEU. We then fine-tune Qwen3-8B~\cite{yang2025qwen3} on MEU-Instruct, equipping the model to interpret these prompts and effectively map quantitative motion data to high-level ME analysis. In essence, the overall MELLM framework grounds high-level reasoning in the precise motion cues captured by MEFlowNet, translating subtle motion patterns into human-readable descriptions with reliable emotional inferences. Our main contributions are as follows:

\begin{itemize}
    \item[$\bullet$]
    This paper marks a transition from the MER paradigm, based on discrete emotion classification, to a MEU framework aimed at comprehensive perception and characterization of subtle facial dynamics and underlying emotional cues. To realize this, we propose a novel ME Large Language Model (MELLM). 

    \item[$\bullet$] We propose MEFlowNet, a warping-based iterative optical flow estimator specifically designed for extracting subtle facial motion features. To facilitate its training and evaluation, we have constructed MEFlowDataset, a large-scale ME optical flow dataset covering diverse identities and subtle facial movements.

    \item[$\bullet$] 
    We introduce a FGMU paradigm, where optical flow signals extracted by MEFlowNet are transformed into structured motion prompts, which are then used to build MEU-Instruct, a specialized instruction-tuning dataset that enables LLMs to better interpret MEs.

    \item[$\bullet$]  
    Extensive experiments demonstrate that MEFlowNet achieves significant improvements in facial and ME flow estimation over existing optical flow methods, and our MELLM attains SOTA accuracy and exhibits strong cross-dataset generalization across multiple ME benchmarks.
\end{itemize}

The remainder of this paper is organized as follows. Section II reviews related work from multiple perspectives. Section III describes our proposed approach, including the task definition and detailed methods for dataset construction and model architecture. Section IV presents the experimental setup, results, and analysis, followed by a case study that demonstrates the practical performance of the proposed method. Finally, Section V concludes the paper and outlines directions for future work.

\section{Related Work}
In this section, we will review and summarize research work related to this paper from three distinct areas: Optical Flow Estimation, Micro-expression Recognition, and Multimodal Large Language Models.
\subsection{Optical Flow Estimation}
Classical optical-flow algorithms such as TV-L1~\cite{werlberger2011optical} and Farnebäck~\cite{farneback2003two} remain popular in MER because they are simple and produce dense motion fields. However, these methods depend on the brightness-constancy assumption and typically presume small inter-frame displacements. In both the CAS(ME)$^3$ dataset~\cite{li2022cas} and real-world recordings, illumination often varies and facial motion can include large rigid components (e.g., head translation and rotation). Under such conditions the brightness-constancy and small-motion assumptions are violated, which can lead classical estimators to produce inaccurate  flows that mask the subtle, localized deformations critical for MER.
Learning-based optical-flow methods relax many of these assumptions and usually improve robustness to large displacements and some illumination changes~\cite{zhai2021optical}. Yet these models are commonly trained and evaluated for overall motion accuracy on standard benchmarks, which does not guarantee sensitivity to the very small, high-frequency facial deformations that define ME.
Lu et al.~\cite{lu2024facialflownet} propose a face-specific method that builds the FacialFlowNet dataset and the DecFlow model to extract facial dynamics and explicitly compensate for head motion. While such approaches reduce the impact of rigid head movement, they still have limited fidelity when it comes to capturing the fine-grained flow patterns produced by MEs.

To address this gap, we introduce the MEFlowDataset and the MEFlowNet model. MEFlowDataset offers more diverse facial appearances and finer-grained facial motion. Furthermore, MEFlowNet is designed to suppress global head motion while better preserving and recovering the subtle flow signals associated with MEs.

\subsection{Micro-expression Recognition} 

As a pivotal modality for decoding human emotional behavior, automatic MER has garnered significant attention in the field of Affective Computing.

Initially, MER research predominantly relied on handcrafted descriptors combined with traditional classifiers~\cite{LBP-TOP, HE201744}. With the advent of deep learning, robust visual representation models, including Convolutional Neural Networks~\cite{zhao2021two} and Vision Transformers~\cite{wang2024jgulf}, have been adopted to extract spatiotemporal features characterizing ME dynamics. Beyond utilizing raw frames or sequences, a prevalent strategy involves computing the optical flow between the onset and apex frames of an ME sequence. This approach effectively mitigates identity-related interference while emphasizing fine-grained motion cues. For instance, FeatRef~\cite{ZHOU2022108275} integrates this strategy into a feature refinement framework, employing attention mechanisms to distill expression-specific features. Similarly, MFDAN~\cite{MFDAN} introduces a dual-branch architecture that leverages optical flow priors to guide the image branch, thereby enabling precise localization of discriminative regions associated with subtle motion patterns.

Notwithstanding the performance gains in discrete classification, existing methods often struggle to capture the granular facial dynamics and emotional nuances inherent in MEs. Furthermore, these approaches frequently suffer from overfitting due to data scarcity and exhibit poor generalization in unconstrained real-world scenarios. To address these limitations, we propose integrating LLMs into MER, harnessing their advanced reasoning capabilities to establish a comprehensive MEU system. Consequently, our proposed MELLM demonstrates superior generalization across diverse recording conditions while preserving high sensitivity to subtle ME dynamics.

\subsection{Multimodal Large Language Models}

In recent years, MLLMs have evolved as prominent tools for various domains and downstream tasks~\cite{Shukang}. Built upon large parameter sizes and large-scale training, MLLMs exhibit unprecedented perception and reasoning capabilities~\cite{JEGHAM2026112765}. These powerful models open up new possibilities and inspire new research paradigms for traditional vision or multimodal tasks. 
For instance, SceneLLM~\cite{ZHANG2026111992} approaches dynamic scene graph generation by transforming video features into language-like tokens, thereby enabling the LLM to perform implicit reasoning on complex spatio-temporal interactions. PLRTQA~\cite{ALAWWAD2025111332} incorporates the retrieval augmented generation technique with LLMs to address the challenges of handling lengthy contexts and reasoning over out-of-domain concepts in textbook question answering.

This trend naturally extends to specialized fields like affective computing. Several MLLMs have been proposed for multimodal emotion understanding by incorporating various combinations of image, video, and audio modalities. For instance, ExpLLM~\cite{lan2024expllm} enables interpretable expression recognition through chain-of-thought reasoning based on AUs, while Emotion-LLaMA~\cite{cheng2024emotion} achieves SOTA results on multimodal emotion benchmarks by fusing audio, visual, and textual information. However, while these models demonstrate strong performance on macro-expressions, their ability to capture the subtle and transient nature of ME remains limited due to insufficient temporal granularity and domain-specific design.

In this work, we combine MEFlowNet’s strength in capturing fine-grained facial dynamics with the outstanding reasoning capability of LLMs, aiming for a more comprehensive and fine-grained understanding of MEs. Different from previous specialist methods, we explore the possibility of adopting a multi-step reasoning process, which aligns with the analysis procedure of human experts and enables a more interpretable method for automatic MEU.

\section{METHODOLOGY}
This section delineates our MEU pipeline. We first define the MEU task and outline the overall framework. Subsequently, we detail the construction of the MEFlowDataset and the architecture of our MEFlowNet model, including its training strategy. We then introduce MEU-Instruct, a specialized instruction-tuning dataset for LLM-based MEU. Finally, we present MELLM, which is fine-tuned and aligned to effectively perform MEU tasks.




\subsection{Task Definition}
In contrast to the conventional MER task, which is confined to classifying ME into discrete emotional categories, we propose an extended formulation termed Micro-Expression Understanding (MEU). This new formulation moves beyond simple classification to foster a more comprehensive and interpretable understanding of these subtle expressions. Specifically, the MEU task comprises three key components:
\begin{itemize}
\item[$\bullet$] Facial Action Analysis: This involves analyzing subtle facial dynamics across various regions to identify and potentially quantify the activation of relevant AUs corresponding to these movements.

\item[$\bullet$] Inferential Reasoning: This component focuses on constructing a logical inference chain from the observed facial dynamics and AU activations to deduce the underlying affective state.

\item[$\bullet$] Conclusion: This step involves providing a summary output containing the identified AUs and inferred ME categories.
\end{itemize}

Therefore, the MEU task is designed to yield an intuitive, justifiable, and transparent interpretation of MEs. By requiring models to incorporate detailed facial analysis and explicit reasoning, this task structure promotes the development of models with enhanced interpretability and generalization capabilities. This, in turn, fosters greater trustworthiness and enables more robust and versatile applications in domains requiring a nuanced understanding of human affective states.

\subsection{Framework Overview}
To advance MEU, we propose a novel framework, MELLM, which integrates fine-grained facial-motion perception with the reasoning capabilities of LLMs, as illustrated in \autoref{fig:figure5}. Our pipeline begins with the construction of MEFlowDataset, a large-scale dataset of synthesized optical-flow sequences tailored for MEs. Next, we introduce MEFlowNet, an optical-flow estimation network trained on MEFlowDataset and explicitly optimized to capture the subtle, fine-grained motion signatures of MEs. Building upon this, we formalize the Flow-Guided Micro-Expression Understanding (FGMU) paradigm. In this paradigm, flow fields extracted by MEFlowNet are transformed into structured motion prompts, which then enable a three-stage analysis performed by an LLM. These prompts form the basis of MEU-Instruct, an instruction-tuning dataset designed to align the LLM with this motion-centric information. Finally, we fine-tune Qwen3-8B on MEU-Instruct, training it to interpret structured motion prompts and map quantitative motion data to high-level ME analysis and inference.

\begin{figure}[t]
  \centering
  \includegraphics[width=1\textwidth]{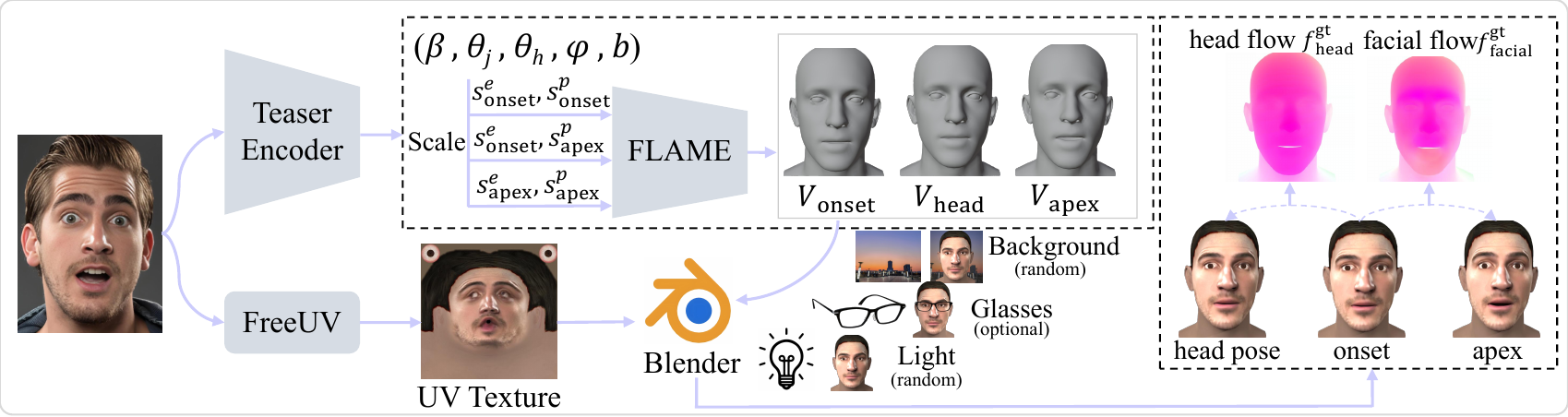}
  \caption{The construction pipeline of the MEFlowDataset. First, FLAME parameters ($\beta, \theta_j, \theta_h, \varphi , b$) and the UV-texture are extracted from the input facial image using the Teaser Encoder and FreeUV, respectively. These parameters are then scaled and fed into the FLAME model to generate three distinct meshes: the onset $V_{\text{onset}}$, head pose $V_{\text{head}}$, and apex $V_{\text{apex}}$ facial meshes. Finally, synthetic facial images and their corresponding ground truth head $f^{\text{gt}}_{\text{head}}$ and facial $f^{\text{gt}}_{\text{facial}}$ flows are generated using Blender.}
  \label{fig:figure3}
\end{figure}

\subsection{Facial Subtle Motion Perception}
\subsubsection{MEFlowDataset}
To provide reliable ground-truth supervision for training our optical flow estimator, MEFlowNet, we present a synthetic ME optical flow dataset named MEFlowDataset. The dataset comprises 54,611 onset-to-apex image pairs across diverse identities, accompanied by corresponding facial-flow and head-flow fields. As depicted in \autoref{fig:figure3}, our data generation pipeline is inspired by FacialFlowNet~\cite{lu2024facialflownet} and leverages source images from AffectNet~\cite{mollahosseini2017affectnet}, a large-scale dataset of real-world facial expressions.

The process begins by feeding an input facial image into the Teaser Encoder~\cite{liu2025teaser}, a 3D expression reconstruction model, to estimate five key facial parameters: shape coefficients $\beta\in\mathbb{R}^{300}$, jaw rotation $\theta_j\in\mathbb{R}^{3}$, head pose $\theta_h\in\mathbb{R}^{3}$, expression coefficients $\varphi\in\mathbb{R}^{50}$, and eyelid closure blendshapes $b\in\mathbb{R}^{2}$. Since the recovered parameters typically represent macro-expressive motion (i.e., large-magnitude changes in facial expression and head pose), we attenuate them by applying scaling factors to synthesize ME dynamics.

Specifically, we generate three distinct meshes—an onset mesh $V_{\mathrm{onset}}$, a head-pose mesh $V_{\mathrm{head}}$, and an apex mesh $V_{\mathrm{apex}}$—by employing the FLAME model~\cite{FLAME:SiggraphAsia2017}. FLAME is a powerful 3D facial model trained on thousands of detailed 3D scans, capable of representing a wide variety of facial shapes and expressions. The meshes are produced by evaluating the FLAME model with scaled parameters as follows:

\begin{align}
V_{\mathrm{onset}} &= \mathrm{FLAME}\big(\beta,\; s^{e}_{\mathrm{onset}}\,\theta_j,\; s^{p}_{\mathrm{onset}}\,\theta_h,\; s^{e}_{\mathrm{onset}}\,\varphi,\; s^{e}_{\mathrm{onset}}\,b\big), \\
V_{\mathrm{head}}  &= \mathrm{FLAME}\big(\beta,\; s^{e}_{\mathrm{onset}}\,\theta_j,\; s^{p}_{\mathrm{apex}}\,\theta_h,\; s^{e}_{\mathrm{onset}}\,\varphi,\; s^{e}_{\mathrm{onset}}\,b\big), \\
V_{\mathrm{apex}}  &= \mathrm{FLAME}\big(\beta,\; s^{e}_{\mathrm{apex}}\,\theta_j,\; s^{p}_{\mathrm{apex}}\,\theta_h,\; s^{e}_{\mathrm{apex}}\,\varphi,\; s^{e}_{\mathrm{apex}}\,b\big),
\end{align}
where $s^{e}_{\mathrm{onset}}$ and $s^{p}_{\mathrm{onset}}$ represent the scaling coefficients for expression-related and head pose parameters applied to generate the onset state of a synthetic ME. Similarly, $s^{e}_{\mathrm{apex}}$ and $s^{p}_{\mathrm{apex}}$ are the corresponding coefficients used to generate the apex state.

We generate UV-textures using FreeUV~\cite{yang2025freeuv}, a framework for recovering realistic facial textures from single 2D images. We further augment the scenes with randomized backgrounds, optionally add eyeglasses, and introduce variations in illumination. The textured meshes are then rendered into photorealistic onset and apex images using Blender's Cycles rendering engine\footnote{\url{https://www.blender.org/}} , from which we derive the corresponding ground truth facial flow $f^{\mathrm{gt}}_{\mathrm{facial}}$ and head flow $f^{\mathrm{gt}}_{\mathrm{head}}$. 

Unlike FacialFlowNet~\cite{lu2024facialflownet}, which contains large-scale movements and a limited set of identities, our MEFlowDataset is tailored for subtle facial dynamics. Our pipeline reconstructs complete 3D faces from single images and assigns a unique identity to each sample, yielding a more diverse and realistic corpus that provides a stronger foundation for training generalizable models.

\subsubsection{MEFlowNet}

\begin{figure}[t]
  \centering
  \includegraphics[width=1\textwidth]{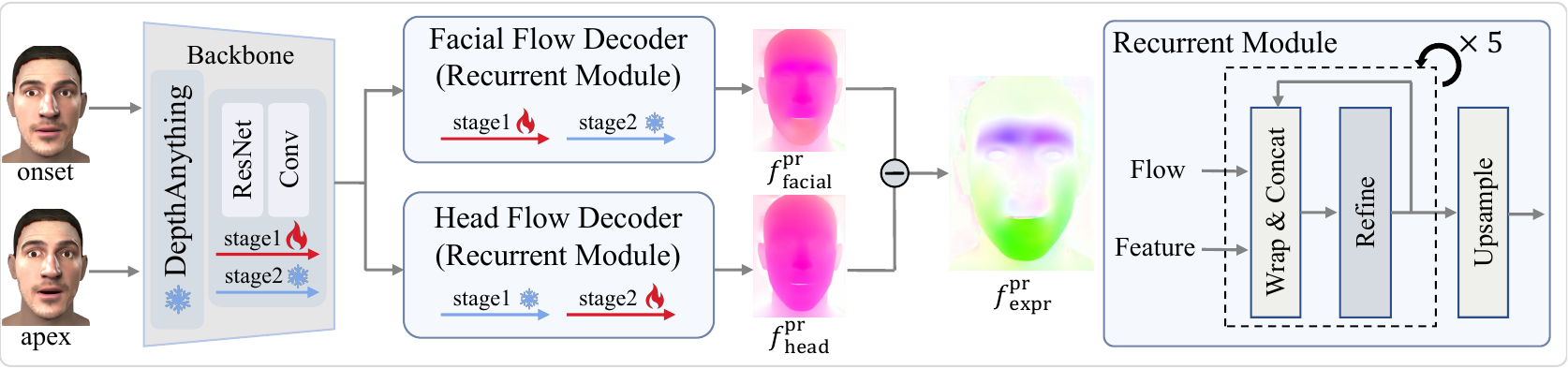}
  \caption{The architecture of MEFlowNet, which employs a two-stage training method. The predicted ME flow $f^{\text{pr}}_{\text{expr}}$ is derived by subtracting the head flow $f^{\text{pr}}_{\text{head}}$ from the facial flow $f^{\text{pr}}_{\text{facial}}$.}
  \label{fig:figure4}
\end{figure}

To capture fine-grained expression signals from global facial motion, we introduce MEFlowNet, which adopts the warping-based iterative architecture proposed in WAFT~\cite{wang2025waft} as its core optical flow extractor. As illustrated in \autoref{fig:figure4}, features are first extracted using a frozen DepthAnythingV2~\cite{yang2024depth} model and subsequently refined by a ResNet module. These features are then upsampled and fused through a series of convolutional layers. We instantiate two parallel recurrent modules, adapted from the DPT~\cite{ranftl2021vision} architecture, to separately estimate the facial flow $f^{\mathrm{pr}}_{\mathrm{facial}}$ and head flow $f^{\mathrm{pr}}_{\mathrm{head}}$. The final ME flow, $f^{\mathrm{pr}}_{\mathrm{expr}}$, is obtained by removing the head flow components from the facial flow.

We employ a two-stage training scheme to disentangle rigid head motion from facial motion. In Stage 1, the model is trained to capture the global facial displacement between onset and apex frames. In Stage 2, we freeze the backbone and the Facial Flow Decoder, training only the Head Flow Decoder to predict head motion. This allows the ME flow to be recovered as the residual. We detail the specifics of each stage below.

\begin{figure}[t]
    \centering
    \includegraphics[width=1\textwidth]{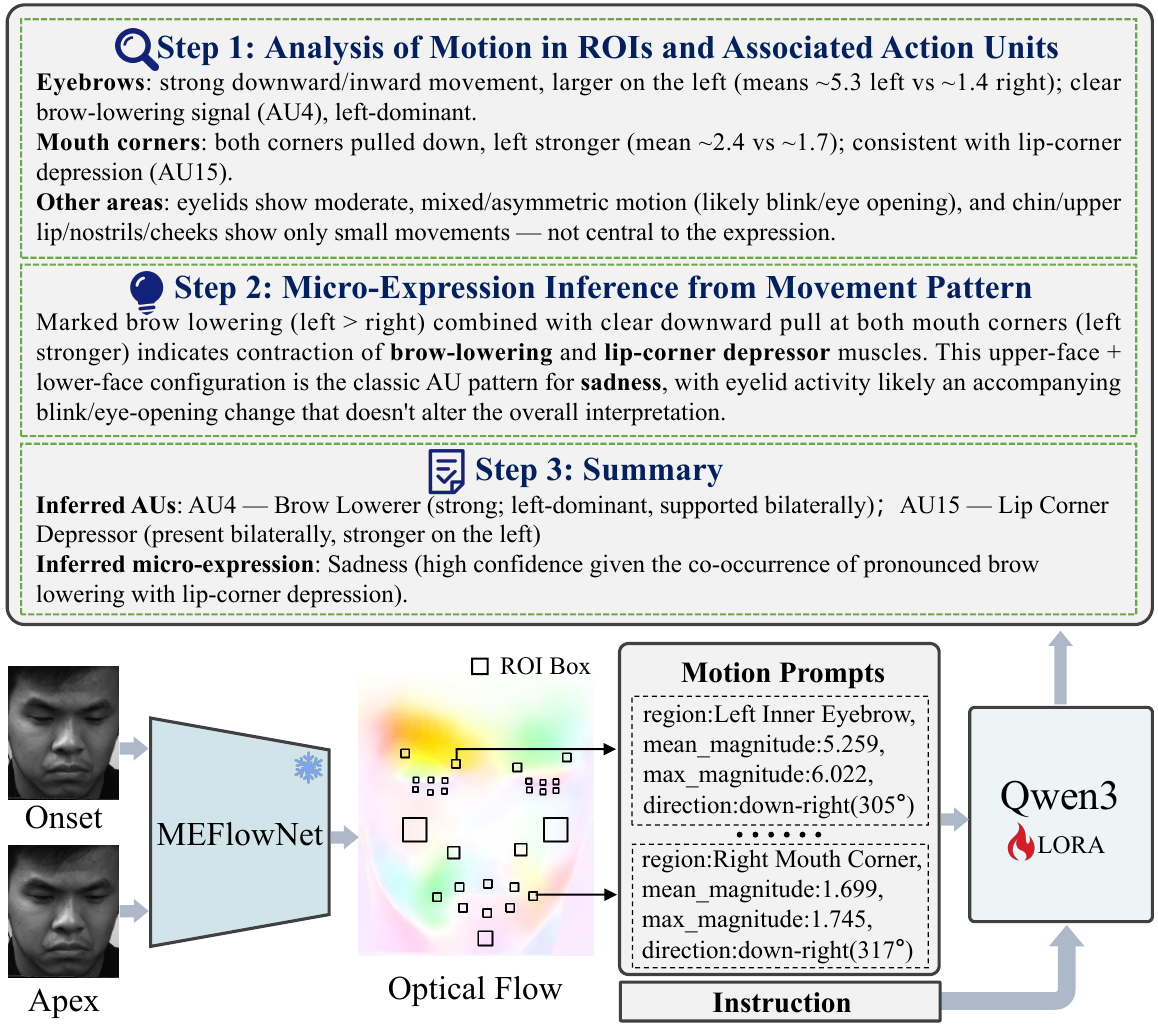}
    \caption{The overall architecture of the MELLM.}
    \label{fig:figure5}
\end{figure}

\textbf{Stage 1.} Both the DepthAnythingV2 module and the Head Flow Decoder are frozen, while the remaining backbone components and the Facial Flow Decoder are trained. The objective in this stage is
\begin{align}
\mathcal{L}_{\text{stage1}} = \mathcal{L}_{\mathrm{MoL}}^{\mathrm{facial}} + \mathcal{L}_{\mathrm{ROI}}^{\mathrm{facial}},
\end{align}
where $\mathcal{L}_{\mathrm{MoL}}^{\mathrm{facial}}$ denotes the Mixture-of-Laplace (MoL) loss as used in~\cite{wang2025waft, wang2024sea}, and the Region of Interest (ROI) endpoint-error term is defined by
\begin{align}
\mathcal{L}_{\mathrm{ROI}}^{\mathrm{facial}} = \mathrm{EPE}\!\left( \big(f_{\mathrm{facial}}^{\mathrm{pr}} - f^{\mathrm{gt}}_{\mathrm{facial}}\big) \odot \mathrm{mask}_{\mathrm{ROI}} \right),
\end{align}
Here, $f_{\mathrm{facial}}^{\mathrm{pr}}$ denotes the facial flow predicted by the Facial Flow Decoder, and $f^{\mathrm{gt}}_{\mathrm{facial}}$ denotes the corresponding ground-truth facial flow. Our ROI comprises 29 manually defined key facial regions that effectively capture facial micro-movements, as shown in \autoref{fig:figure5}. The term $\mathrm{mask}_{\mathrm{ROI}}$ is the binary spatial mask derived from these regions, where pixels inside the ROI are set to a value of 1 and all pixels outside are set to 0. The symbol $\odot$ denotes element-wise multiplication. Prior to this multiplication, $\mathrm{mask}_{\mathrm{ROI}}$ is broadcast to match the number of flow channels.

\textbf{Stage 2.} We optimize the Head Flow Decoder while keeping the backbone and the Facial Flow Decoder frozen. Define the ground truth $f^{\mathrm{gt}}_{\mathrm{expr}}=f^{\mathrm{gt}}_{\mathrm{facial}}-f^{\mathrm{gt}}_{\mathrm{head}}$. The expression-focused losses are
\begin{align}
\mathcal{L}_{\mathrm{ROI}}^{\mathrm{expr}} = \mathrm{EPE}\!\left( \big(f^{\mathrm{pr}}_{\mathrm{expr}} - f^{\mathrm{gt}}_{\mathrm{expr}}\big) \odot \mathrm{mask}_{\mathrm{ROI}} \right), 
\end{align}

\begin{align}
\mathcal{L}_{\mathrm{expr}} = \mathrm{EPE}\!\left( f^{\mathrm{pr}}_{\mathrm{expr}} - f^{\mathrm{gt}}_{\mathrm{expr}} \right),
\end{align}
The combined objective for Stage 2 is
\begin{align}
\mathcal{L}_{\text{stage2}} = \alpha \cdot \mathcal{L}_{\mathrm{MoL}}^{\mathrm{head}} + \mathcal{L}_{\mathrm{ROI}}^{\mathrm{expr}} + \mathcal{L}_{\mathrm{expr}},
\end{align}
where $\mathcal{L}_{\mathrm{MoL}}^{\mathrm{head}}$ is the MoL loss applied to the head-flow prediction and $\alpha$ is a balancing weight set to 0.5.


\subsection{Flow-Guided Micro-Expression Understanding~(FGMU)}
\subsubsection{The FGMU Paradigm}
In this section, we introduce the FGMU paradigm, which bridges low-level motion features with high-level affective interpretation using optical-flow cues. The process commences by extracting a facial motion flow field from an onset-apex ME pair and identifying key ROIs with facial landmarks~\cite{hu2025openface}. Subsequently, the extracted numerical flow fields are converted into structured motion prompts comprehensible to a LLM. Specifically, for the optical flow within each ROI, we compute three key metrics: its mean magnitude, maximum magnitude, and dominant direction. The direction is then quantized into one of eight discrete vectors and specified by its precise angle. This information is formatted into a textual key-value structure, for example:
\begin{center}
\texttt{\{region: "Left Inner Eyebrow", mean\_magnitude: 5.259, max\_magnitude: 6.022, direction: "down-right (305°)"\}}
\end{center}

The information gathered from all ROIs is aggregated into a structured motion prompt, which is appended with an instruction and serves as the direct input for the LLM. The instruction requires the LLM to perform the following hierarchical three-step analytical procedure:
\textit{1) Analysis of Motion in ROIs and Associated Action Units.} The model analyzes the magnitude and direction of motion within predefined facial ROIs (e.g., eyebrows, eyelids, cheeks, mouth, chin) and, from these localized motion signatures, identifies the corresponding active facial AUs.
\textit{2) Micro-Expression Inference from Movement Pattern.} The model integrates validated AUs with global and local motion patterns to infer the most likely ME category. This inference is guided by established AU-emotion mappings and physiological priors to ensure semantic consistency.
\textit{3) Summary.} The model produces a concise, interpretable report that lists the recognized AUs, the inferred ME category, and the supporting motion evidence.

\subsubsection{Instruction-tuning MEU Dataset}
Guided by the FGMU paradigm, we construct an instruction-tuning dataset, MEU-Instruct, based on the DFME dataset~\cite{zhao2023dfme}, to align facial-motion signals with LLMs. Specifically, we apply MEFlowNet to extract ME flow from each onset-apex image pair and subtract the nasal-tip flow to compensate for residual head motion. From the resulting flow fields, we compute ROI-level facial-motion signals and formatted them into structured motion prompts.

Subsequently, we employ Qwen3-MAX~\cite{yang2025qwen3} to perform a three-stage analysis of the extracted signals in accordance with the FGMU paradigm:

\textbf{Step 1}: Analyze ROIs significantly related to ME and their corresponding AUs.

\textbf{Step 2}: Infer the ME category based on the overall motion pattern.

\textbf{Step 3}: Summarize the AUs and the corresponding ME.

To ensure the fidelity of the generated data, we enforce that Qwen3-MAX's reasoning remains consistent with the ground-truth AU and emotion annotations from the DFME dataset. Acknowledging that the Facial Action Coding System~\cite{ekman1978facial} describes all facial movements, including those unrelated to emotional expression (e.g., eye blinks), our process handles these as well. Qwen3-MAX is prompted to analyze the motion signals for these non-expressive movements but is explicitly instructed not to assign them AU labels if they are absent from the ground-truth annotation. This meticulous process establishes a transparent reasoning chain—from low-level optical-flow features, through AU-level analysis, to expression-category inference—thereby enhancing the semantic grounding and reliability of MEU-Instruct dataset.

\subsubsection{MELLM Framework and Adaptation for MEU}
The overall architecture of MELLM is depicted in \autoref{fig:figure5}. The model comprises two core components: MEFlowNet, an optical flow estimator, and Qwen3-8B~\cite{yang2025qwen3}, an LLM fine-tuned for MEU using our MEU-Instruct dataset.

The inference process begins with an onset-apex frame pair, from which MEFlowNet computes an optical-flow field. From this field, ROI-level motion signals detailing intensity and dominant direction are extracted. These signals are then formatted into a structured motion prompt, which is combined with a task-specific instruction and fed into the fine-tuned LLM.

To analyze ME under the FGMU paradigm, we fine-tune Qwen3-8B on the MEU-Instruct dataset. We employ Low-Rank Adaptation (LoRA) for parameter-efficient fine-tuning, updating only a small set of lightweight modules. This strategy enables the model to effectively interpret subtle facial cues and infer the corresponding MEs, all while minimizing the number of trainable parameters.


\section{EXPERIMENTS}
In this section, we present the experimental settings, introduce the datasets along with the evaluation protocols and metrics, and conduct a comprehensive analysis of our proposed methods, concluding with a case study.

\subsection{Experimental Settings}
For MEFlowDataset synthesis, the scaling parameters for the onset mesh, denoted as \(s^{e}_{\mathrm{onset}}\) (expression) and \(s^{p}_{\mathrm{onset}}\) (pose), are uniformly sampled from the interval \([0,\,0.3]\). The expression scaling at the apex mesh, \(s^{e}_{\mathrm{apex}}\), is derived by adding an increment sampled from \([0.03,\,0.25]\) to \(s^{e}_{\mathrm{onset}}\), with 85\% of these increments constrained to values smaller than 0.1.  Similarly, the pose scaling at the apex mesh, \(s^{p}_{\mathrm{apex}}\), is obtained by adding an increment sampled from \([0.0,\,0.1]\) to \(s^{p}_{\mathrm{onset}}\), where 95\% of the increments are less than 0.03. These settings are designed to balance fine-grained facial-motion variation with dataset robustness. Furthermore, to enhance motion-scale diversity during MEFlowNet training, we additionally incorporate a subset of the FacialFlowNet dataset consisting of samples in which facial motion has been downscaled by a factor of 20. MEFlowNet is trained for 55k steps in both Stage~1 and Stage~2, with a batch size of 16 and a learning rate of \(5\times10^{-5}\). The model is initialized using weights from WAFT~\cite{wang2025waft}, which was pre-trained for 200k steps on the \textit{Spring(train)}\cite{mehl2023spring} subset.

From the DFME dataset, we select 2,681 samples for the 3-class task and 3,631 samples for the 7-class classification task. These samples are used with Qwen3-MAX to generate instruction-tuning data for MEU-Instruct. We then fine-tune MELLM for 5 epochs on each task, based on the Qwen3-8B model, employing LoRA for parameter-efficient tuning. LoRA adapters are integrated into the model, with only these adapters are being updated during fine-tuning. The LoRA configuration using a rank \(r=8\) and a scaling parameter \(\alpha=16\). For optimization, we employ the AdamW optimizer with a learning rate of \(1\times10^{-4}\) and a cosine learning rate scheduler.





\subsection{Evaluation Datasets}
We evaluate the MEFlowNet model on the MEFlowDataset and FacialFlowNet test sets. Specifically, we use all 5,461 samples from the MEFlowDataset test set. From the FacialFlowNet test set, we select a subset of 4,446 samples, where the facial motion was downsampled to 1/20 of its original rate.

To assess the effectiveness and generalization capability of MELLM, we conduct a comprehensive evaluation across four distinct ME datasets.
Our primary evaluation focuses on generalization to unseen data, for which we employ a zero-shot testing protocol on three widely-used benchmarks: CASME II~\cite{yan2014casme}, CAS(ME)$^3$~\cite{li2022cas}, and SAMM~\cite{davison2016samm}. This challenging cross-dataset scheme is crucial for assessing the model's robustness and its potential for practical, real-world applications where data will inevitably differ from the training domain.

\textbf{CASME~II}: The CASME II dataset contains 247 ME clips from 26 subjects, recorded at 200 fps with a facial resolution of $280\times340$ pixels. Adopting a three-class labeling scheme, we select 129 samples categorized into Positive (happiness), Negative (sadness, fear, disgust), and Surprise.

\textbf{CAS(ME)$^3$}: We use Part A of the CAS(ME)$^3$ dataset, which is partitioned by elicitation method and contains 860 ME samples. From this subset, we select 699 samples belonging to the three target categories: Positive (happiness), Negative (sadness, fear, anger, disgust), and Surprise.

\textbf{SAMM}: The SAMM dataset consists of 159 ME clips from 32 participants of 13 ethnic backgrounds, recorded at 200 fps with a $2040\times1088$ pixel resolution. For our evaluation, we use 121 samples classified into Positive (happiness), Negative (sadness, fear, anger, disgust), and Surprise.

We also assess the model's foundational performance within its source domain. Since our training data, MEU-Instruct, is generated from the \textbf{DFME} dataset~\cite{zhao2024dynamic}, we evaluate MELLM on its TestA (474 samples) and TestB (299 samples) splits. These test sets cover seven basic emotion labels: happiness, disgust, contempt, surprise, fear, anger, and sadness.

\subsection{Evaluation Protocol and Metrics}

For the optical-flow model, we adopt the End-Point Error (EPE) as the evuation metric. The EPE for a single test sample is defined as:
\begin{equation}
\text{EPE} \;=\; \frac{1}{N}\sum_{i=1}^{N} \left\lVert \mathbf{f}_i - \widehat{\mathbf{f}}_i \right\rVert_2
\;=\; \frac{1}{N}\sum_{i=1}^{N} \sqrt{(u_i - \hat{u}_i)^2 + (v_i - \hat{v}_i)^2},
\label{eq:epe}
\end{equation}
where \(N\) denotes the total number of evaluated flow vectors, \(\mathbf{f}_i=(u_i,v_i)\) represents the ground-truth flow vector at pixel \(i\), and \(\widehat{\mathbf{f}}_i=(\hat{u}_i,\hat{v}_i)\) is the corresponding estimated flow. We report the average EPE across the entire test set.

During MELLM inference, the predicted emotion category is extracted from the summary section of the generated text and serves as the final output. This prediction is then compared with ground-truth labels to compute quantitative metrics. Adhering to the evaluation protocols outlined in~\cite{zhao2023dfme}, we evaluate model performance using Accuracy (ACC), Unweighted F1-score (UF1), and Unweighted Average Recall (UAR). Specifically, UF1 and UAR are adopted to ensure a balanced evaluation across emotion categories, as these metrics are robust to the class imbalance commonly observed in ME datasets.










\textbf{UAR} represents the macro-averaged recall across all classes, computed as the arithmetic mean of the per-class recall values:

\begin{equation}
\text{UAR} = \frac{1}{C} \sum_{i=1}^{C} \frac{TP_i}{TP_i + FN_i},
\end{equation}
where \( C \) denotes the total number of classes, \( TP_i \) is the number of true positives for class \( i \), and \( FN_i \) is the number of false negatives for class \( i \). 

\textbf{UF1} is defined as the macro-average of the F1-scores computed independently for each class, without weighting by class frequency:

\begin{equation}
\text{UF1} = \frac{1}{C} \sum_{i=1}^{C} \frac{2 \cdot P_i \cdot R_i}{P_i + R_i},
\end{equation}
where \( P_i = \frac{TP_i}{TP_i + FP_i} \) is the precision for class \( i \), \( R_i = \frac{TP_i}{TP_i + FN_i} \) is the recall for class \( i \), and \( FP_i \) denotes the number of false positives for class \( i \).

\subsection{Experiment Results and Analysis}

\begin{figure}[t]
\centering
\begin{minipage}[t]{0.4\linewidth}
    \vspace{0pt} 
    \centering
    \includegraphics[width=\linewidth]{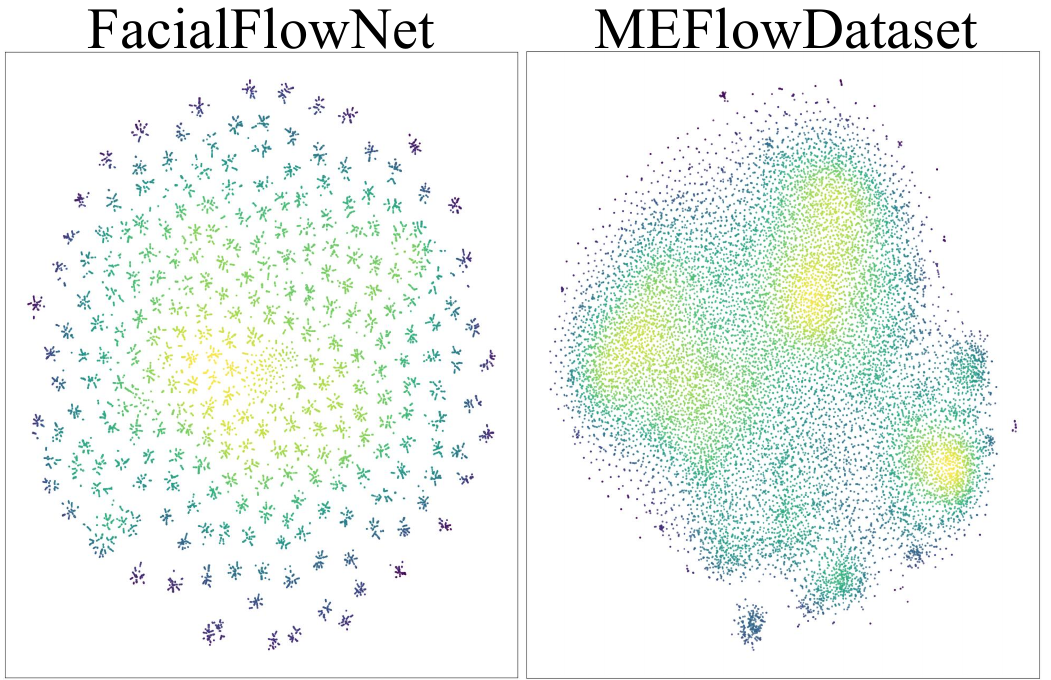} 
    \captionof{figure}{t-SNE visualization of identity features from the FacialFlowNet and MEFlowDataset, based on 20k randomly sampled instances from each dataset.}
    \label{fig:figure6}
\end{minipage}%
\hfill
\begin{minipage}[t]{0.56\linewidth}
    \vspace{0pt} 
    \centering
    \scriptsize
    \captionof{table}{A quantitative comparison of identity diversity, computed on the identity features of 20k randomly sampled instances from each dataset. $\text{std}_{\text{global}}$ is the global mean standard deviation; $\text{cv}_{\text{global}}$ is the global mean coefficient of variation; and $\text{sim}_{\text{std}}$ is the similarity standard deviation. For all metrics, a higher value (indicated by $\uparrow$) signifies greater identity diversity.}
    \begin{tabular}{lcc}
    \toprule
    \textbf{Metrics} & \textbf{FacialFlowNet} & \textbf{MEFlowDataset}\\
    \midrule
    $\text{std}_{\text{global}} \uparrow $ & 0.6998 & \textbf{0.8238}\\
    $\text{cv}_{\text{global}} \uparrow $  & 8.0085 & \textbf{8.5061}\\
    $\text{sim}_{\text{std}} \uparrow $    & 45.270 & \textbf{49.473}\\
    \bottomrule
    \end{tabular}
    \label{tab:table1}
\end{minipage}
\end{figure}

\subsubsection{Analysis of Identity Diversity in MEFlowDataset}
The diversity of an ME optical flow dataset is critical for enabling optical flow models to robustly capture subtle facial features. To evaluate the diversity of our proposed MEFlowDataset, we conducted a comparative analysis of the identity embedding spaces of MEFlowDataset and FacialFlowNet (a facial expression optical-flow dataset). To ensure a fair comparison, we randomly sampled 20,000 images from each dataset and extracted 512-dimensional identity embeddings using ArcFace \cite{deng2019arcface}.

We first visualized these embeddings using t-SNE, as shown in \autoref{fig:figure6}. The visualization indicates that embeddings from FacialFlowNet form many distinct clusters, suggesting limited identity diversity, whereas embeddings from MEFlowDataset lie on a single, continuous manifold, indicating a richer and more continuous identity space. This qualitative observation is corroborated by our quantitative analysis in Table \ref{tab:table1}: MEFlowDataset outperforms FacialFlowNet across all metrics, achieving a $17.7\%$ higher global standard deviation ($\text{std}_{\text{global}}$). Both qualitative and quantitative analyses corroborate the enhanced identity diversity of MEFlowDataset, highlighting our data generation pipeline's ability to create rich data by synthesizing MEs from any single facial image.

\subsubsection{Experiments on optical flow estimation for facial dynamics}
\begin{figure}[t]
    \centering
    \includegraphics[width=1\textwidth]{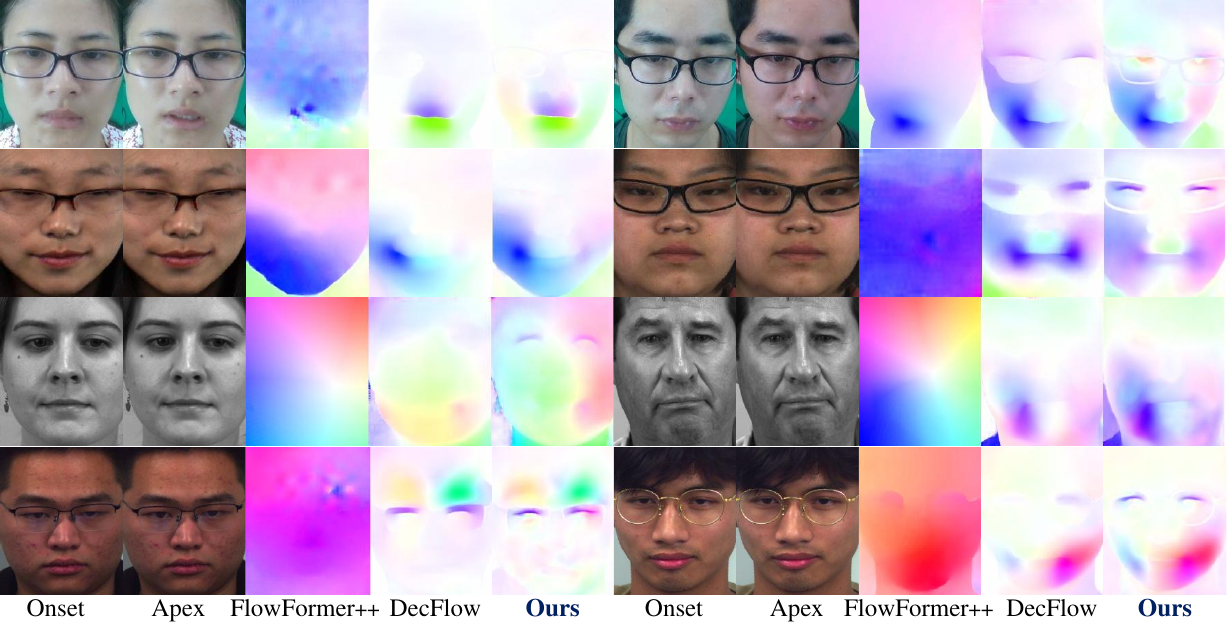}
    \caption{Visualization of optical flow extracted by different methods.}
    \label{fig:figure7}
\end{figure}
As shown in \autoref{tab:table2}, we evaluated the proposed MEFlowNet against three baseline methods: RAFT, FlowFormer++, and the expression-flow method DecFlow. The evaluation was conducted on the FacialFlowNet subset (motion scale = 1/20) and our MEFlowDataset test set. Without fine-tuning, both RAFT and FlowFormer++ demonstrated limited suitability for estimating facial optical flow, whereas DecFlow produced reasonable performance on FacialFlowNet but exhibited a significant decline on the more challenging MEFlowDataset.
Fine-tuning all three baselines on our training data led to substantial performance gains. Despite these improvements, MEFlowNet consistently achieved superior results across all evaluation settings. In particular, for ME flow prediction on the MEFlowDataset test set, MEFlowNet attained an EPE of 0.103, which corresponds to a relative error reduction of $\approx 73.25\%$ compared to the EPE of 0.385 obtained by DecFlow.

This quantitative superiority is further corroborated by our qualitative analysis. As visualized in \autoref{fig:figure7}, MEFlowNet exhibits significantly better robustness and generalization compared to FlowFormer++ and DecFlow. It more effectively captures the subtle facial motions in realistic scenarios of MEs, whereas the baselines tend to produce noisy or incomplete flow fields.

\begin{table}[]
\centering
\caption{Quantitative comparison of flow-estimation models measured by EPE. For the FacialFlowNet test set, we selected the subset with the motion scale of 1/20. “*” indicates models fine-tuned on the same data used to train MEFlowNet.}
\scriptsize

\begin{tabular}{lcccc}
\toprule 
\multirow{2}{*}{Models} & \multicolumn{2}{c}{Facial Flow Estimation} & \multicolumn{2}{c}{ME Flow Estimation} \\ 

\cmidrule(lr){2-3} \cmidrule(lr){4-5} 
 & FacialFlowNet & MEFlowDataset & FacialFlowNet & MEFlowDataset \\ 
\midrule 

RAFT\cite{teed2020raft}             & 0.275 & 0.280 & / & / \\
FlowFormer++\cite{shi2023flowformer++}     & 0.278 & 0.301 & / & / \\
DecFlow\cite{lu2024facialflownet}          & 0.098 & 0.272 & 0.098 & 0.385 \\ 
\midrule 

RAFT*           & 0.093 & 0.123 & / & / \\
FlowFormer++*   & 0.078 & 0.106 & / & / \\
DecFlow*        & 0.088 & 0.123 & 0.083 & 0.132 \\ 
\midrule 

MEFlowNet (Ours) & \textbf{0.078} & \textbf{0.094} & \textbf{0.076} & \textbf{0.103} \\ 
\bottomrule 
\end{tabular}
\label{tab:table2}
\end{table}

\subsubsection{Experiments on ME recognition and understanding}
\textbf{Performance of Expert Models.}
To establish a baseline, we reproduced several SOTA, task-specific MER models and evaluated them under consistent training and testing protocols. While these expert models exhibit high performance when trained and tested within the same dataset, they exhibit limited generalization capability across different datasets. To assess this limitation systematically, we trained these models on our DFME dataset, using the same training samples as MELLM, and evaluated their performance on three external benchmark datasets: CASME II, CAS(ME)$^3$, and SAMM. As summarized in \autoref{tab:table3}, most expert models experience a substantial performance drop in this cross-dataset setting. Although MMNet demonstrates relatively better robustness, the overall results reveal a pronounced overfitting issue. These findings suggest that existing specialized MER classifiers are heavily optimized for particular datasets, which are often limited in scale and diversity, thereby constraining their applicability to more varied and in-the-wild scenarios.

\begin{table}[ht]
\centering
\caption{Comparison of zero-shot three-class classification performance with SOTA methods on the CASME II, CAS(ME)$^3$, and SAMM datasets.}
\label{tab:table3}
\resizebox{\textwidth}{!}{%
\begin{tabular}{llcccccccc}
\toprule
\multirow{2}{*}{\textbf{Models}} & \multirow{2}{*}{\textbf{Methods}} & \multicolumn{2}{c}{\textbf{CASME II}} & \multicolumn{2}{c}{\textbf{CAS(ME)$^3$}} & \multicolumn{2}{c}{\textbf{SAMM}} & \multicolumn{2}{c}{\textbf{Average}} \\
\cmidrule(lr){3-4} \cmidrule(lr){5-6} \cmidrule(lr){7-8} \cmidrule(lr){9-10}
 & & UF1 & UAR & UF1 & UAR & UF1 & UAR & UF1 & UAR \\
\midrule

LBP-TOP\cite{pfister2011recognising} & \multirow{5}{*}{Expert Model} & 0.3282 & 0.4726 & 0.3183 & 0.3962 & 0.2720 & 0.3368 & 0.3062 & 0.4019 \\
STSTNet\cite{liong2019shallow}       & & 0.1838 & 0.1904 & 0.2403 & 0.2407 & 0.2404 & 0.2377 & 0.2215 & 0.2229 \\
Capsule\cite{van2019capsulenet}      & & 0.3353 & 0.4031 & 0.4365 & \uline{0.4568} & 0.4144 & 0.4372 & 0.3954 & 0.4324 \\
MMNet\cite{li2022mmnet}              & & \uline{0.4802} & \uline{0.4833} & \uline{0.4591} & 0.4526 & \uline{0.4736} & \uline{0.4557} & \uline{0.4710} & \uline{0.4639} \\
HTNet\cite{wang2024htnet}            & & 0.1706 & 0.1690 & 0.2833 & 0.2878 & 0.3384 & 0.3491 & 0.2641 & 0.2686 \\
\midrule 

InternVL3.5~14B\cite{wang2025internvl3} & \multirow{2}{*}{MLLM} & 0.2245 & 0.1499 & 0.1886 & 0.1323 & 0.1493 & 0.0933 & 0.1875 & 0.1252 \\
Qwen2.5-VL~32B\cite{bai2025qwen2}       & & 0.3162 & 0.2975 & 0.2761 & 0.2592 & 0.3648 & 0.3366 & 0.3190 & 0.2978 \\
\midrule 

MELLM (Ours) & FGMU paradigm & \textbf{0.8261} & \textbf{0.8524} & \textbf{0.5681} & \textbf{0.6170} & \textbf{0.5573} & \textbf{0.6045} & \textbf{0.6505} & \textbf{0.6913} \\
\bottomrule
\end{tabular}%
}
\end{table}

\begin{table}[t]
\centering
\caption{Experimental Results Compared to general MER methods on DFME Dataset (TESTA \& TESTB) for seven-class classification.}
\label{tab:table4}
\scriptsize
\begin{tabular}{llccc}
\toprule 
\textbf{MER Methods} & \textbf{Test Set} & \textbf{UF1} & \textbf{UAR} & \textbf{ACC} \\ 
\midrule 

FeatRef\cite{ZHOU2022108275}      & \multirow{4}{*}{TestA} & 0.3410 & 0.3686 & \textbf{0.5084} \\
Wang et al.\cite{zhao2024dynamic} &                        & 0.4067 & 0.4074 & 0.4641 \\
He et al.\cite{zhao2024dynamic}~  &                        & \textbf{0.4123} & \textbf{0.4210} & 0.4873 \\
MELLM (Ours)                      &                        & 0.3630 & 0.3788 & 0.4747 \\ 

\midrule 

FeatRef\cite{ZHOU2022108275}      & \multirow{4}{*}{TestB} & 0.2875 & 0.3228 & 0.3645 \\
Wang et al.\cite{zhao2024dynamic} &                        & 0.3534 & 0.3661 & 0.3813 \\
He et al.\cite{zhao2024dynamic}~  &                        & \textbf{0.4016} & \textbf{0.4008} & \textbf{0.4147} \\
MELLM (Ours)                      &                        & 0.3410 & 0.3538 & 0.3779 \\ 
\bottomrule 
\end{tabular}
\end{table}

\textbf{Performance of MLLMs.}
We then investigated the capabilities of general-purpose MLLMs for MER. Due to privacy constraints of the test datasets, we locally deployed two prominent open-source MLLMs, InternVL-3.5 (14B) and Qwen-2.5-VL (32B), for evaluation. Each sample, consisting of an onset and an apex frame, was presented to the models with the following prompt:
"Identify the dynamic changes in facial features between two images, recognize the Action Units, and infer the Micro-Expression being conveyed. Finally, classify the Micro-Expression into one of the following categories: positive, negative, or surprise, and clearly state the chosen category in the answer."
Since these MLLMs generate free-form text rather than standardized labels, we employed the Qwen3-Plus API to post-process their responses and extract the final emotion classifications. The evaluation results, summarized in \autoref{tab:table3}, reveal notable limitations. A substantial portion of the models' outputs failed to identify any relevant facial dynamics or expressions—$69.65\%$ for InternVL-3.5 and $25.08\%$ for Qwen-2.5-VL. As a result, both models exhibited poor overall performance on the MEU task. These findings indicate that general-purpose MLLMs, in their current off-the-shelf form, remain inadequate for handling the nuanced and fine-grained demands of MEU.

\textbf{Performance of MELLM.}
In contrast, our proposed MELLM demonstrates superior cross-dataset generalization. It outperforms the strongest baseline, MMNet, by substantial margins, achieving relative improvements of $38.10\%$ in average UF1 and $49.02\%$ in average UAR across the three external datasets. Furthermore, on the in-domain DFME test sets (TestA and TestB), MELLM also surpasses the FeatRef baseline in both UF1 and UAR, as detailed in \autoref{tab:table4}. It is worth noting that MELLM's performance is slightly lower than that of the top-performing specialized classifiers reported in~\cite{zhao2024dynamic}, which are tailored for this task through end-to-end optimization and task-specific regularization. This performance gap could be associated with the distinct learning paradigms. Unlike traditional expert models that are narrowly optimized for ME classification and tend to overfit on limited data, MELLM employs a perception-and-reasoning pipeline that performs fine-grained motion perception and leverages an LLM for emotion inference. This approach fosters a more holistic understanding of facial dynamics in context, thereby substantially enhancing cross-domain robustness. The improved generalization, however, comes at the cost of a modest reduction in raw classification metrics on the in-domain test sets.

\begin{table}[]
    \centering
    \caption{Ablation Study on LLM models and optical flow methods.}
    \label{tab:table5}
    \renewcommand{\arraystretch}{1.2} 
    \resizebox{\textwidth}{!}{%
        \begin{tabular}{llcccccccc}
            \toprule 
            
            \multirow{2}{*}{\textbf{LLM Models}} &
            \multirow{2}{*}{\textbf{\makecell[l]{Optical Flow\\Methods}}} &
            \multicolumn{2}{c}{\textbf{CASME\textsc{II}}} &
            \multicolumn{2}{c}{\textbf{CAS(ME)$^3$}} &
            \multicolumn{2}{c}{\textbf{SAMM}} &
            \multicolumn{2}{c}{\textbf{Average}} \\ 
            
            \cmidrule(lr){3-4} \cmidrule(lr){5-6} \cmidrule(lr){7-8} \cmidrule(lr){9-10}
            
            & & UF1 & UAR & UF1 & UAR & UF1 & UAR & UF1 & UAR \\ 
            \midrule 
            
            Qwen3-8B\cite{yang2025qwen3} &
            \multirow{2}{*}{MEFlowNet (Ours)} &
            0.2570 & 0.3813 &
            0.3429 & 0.4117 &
            0.1801 & 0.2975 &
            0.2600 & 0.3635 \\
            
            Qwen3-MAX\cite{yang2025qwen3} & &
            0.5960 & 0.5696 &
            0.4880 & 0.4824 &
            \underline{0.5488} & \underline{0.5459} &
            0.5443 & 0.5326 \\ 
            \midrule 
            
            \multirow{4}{*}{\makecell[l]{Qwen3-8B Finetuned \\ w/ MEU-Instruct \\ (Ours)}} &
            TV-L1\cite{werlberger2011optical} &
            0.5792 & 0.6313 &
            \underline{0.5580} & 0.5932 &
            0.4398 & 0.4849 &
            0.5257 & 0.5698 \\
            
            & FlowFormer++\cite{shi2023flowformer++} &
            0.4616 & 0.5404 &
            0.5561 & 0.5779 &
            0.4070 & 0.4676 &
            0.4749 & 0.5286 \\
            
            & DecFlow\cite{lu2024facialflownet} &
            \underline{0.6279} & \underline{0.6533} &
            0.5516 & \underline{0.5962} &
            0.5000 & 0.5261 &
            \underline{0.5598} & \underline{0.5919} \\
            
            & MEFlowNet (Ours) &
            \textbf{0.8261} & \textbf{0.8524} &
            \textbf{0.5681} & \textbf{0.6170} &
            \textbf{0.5573} & \textbf{0.6045} &
            \textbf{0.6505} & \textbf{0.6913} \\ 
            \bottomrule 
        \end{tabular}%
    }
\end{table}

\subsection{Ablation Study}

We conducted a series of ablation studies to systematically evaluate the individual contributions of the two core components in our proposed framework: the LLM backbone and the optical-flow estimator.


\textbf{Impact of LLM Scale and Fine-Tuning.} In the first experiment, we evaluated the contribution of the LLM by keeping the optical flow estimator fixed as MEFlowNet. We explored three LLM configurations:
(a) \textit{Qwen3-8B}, a small pre-trained model without fine-tuning;
(b) \textit{Qwen3-MAX}, a more powerful pre-trained model without fine-tuning; and
(c) our fine-tuned \textit{Qwen3-8B} model trained on the MEU-Instruct dataset. 

The results, summarized in \autoref{tab:table5}, reveal a clear performance hierarchy.  
The base Qwen3-8B, representative of small-scale LLMs, struggled to infer ME from motion cues. Qwen3-MAX, owing to its stronger reasoning capability, achieved substantially better inference and prediction performance. Most notably, our fine-tuned Qwen3-8B not only closed the performance gap but also surpassed Qwen3-MAX.

\textbf{Efficacy of the Optical-Flow Estimator.}
In the second experiment, we evaluated the impact of the optical flow component while fixing the LLM to our fine-tuned Qwen3-8B model. We compared MEFlowNet against three representative optical flow estimators: TV-L1, FlowFormer++, and DecFlow. To ensure a fair comparison, all methods were applied with consistent nose-tip-based head motion correction. As shown in \autoref{tab:table5}, MEFlowNet consistently outperformed the other estimators. Although DecFlow, which is specifically designed for facial analysis, and TV-L1, known for capturing fine-grained motion, both produced competitive results, MEFlowNet demonstrated superior performance. This advantage can be attributed to its synergistic design, which integrates active head-motion compensation, enhanced robustness in complex scenarios, and increased sensitivity to the subtle facial dynamics characteristic of ME.

\begin{figure}[!htbp]
  \centering
  \includegraphics[width=\linewidth]{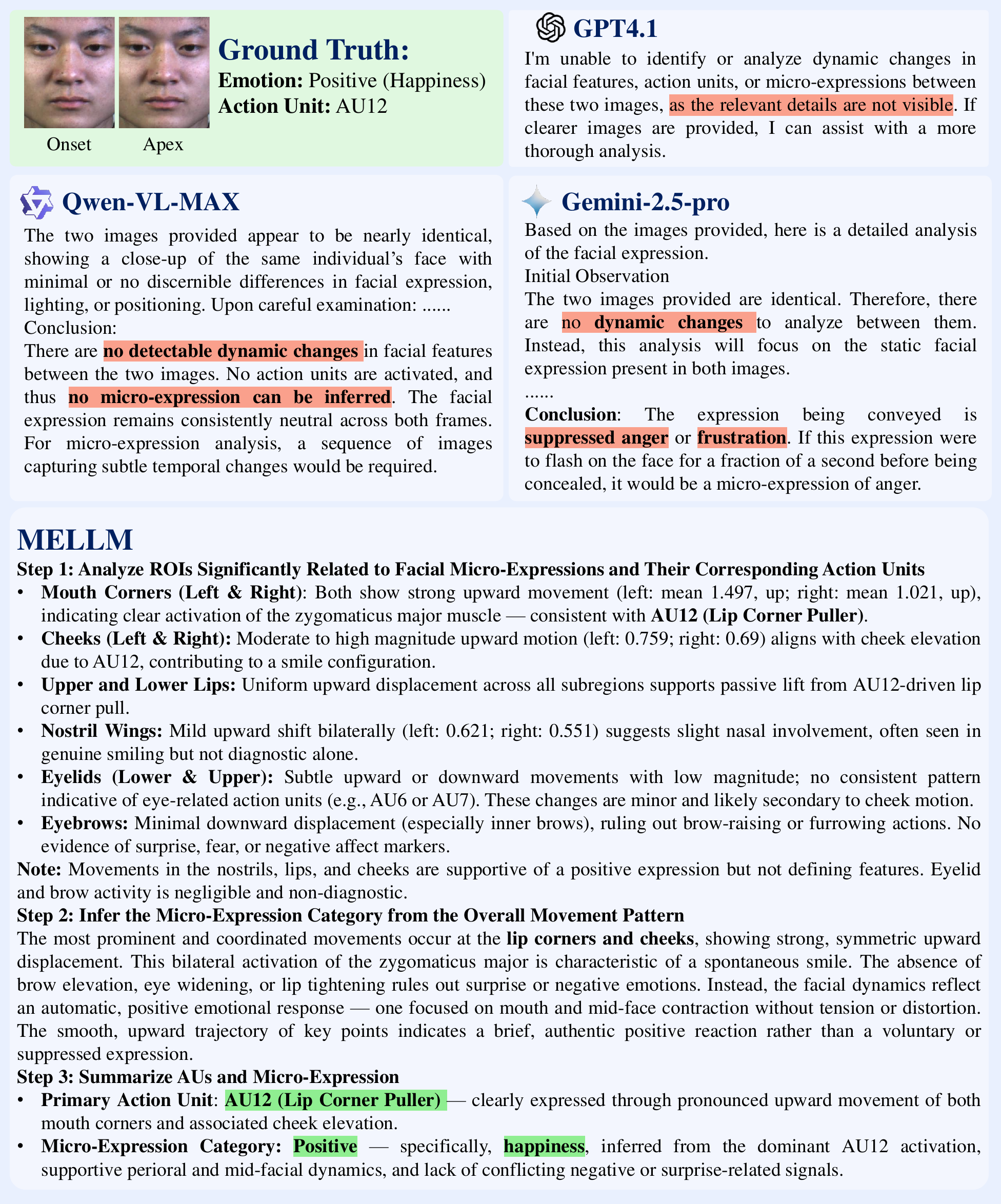}
  \caption{An example comparing MELLM with GPT4.1, Gemini-2.5-pro and Qwen-VL-Max in MEU. Correct predictions are highlighted with a green background, while incorrect ones are highlighted in red.}
  \label{fig:figure8}
\end{figure}

\subsection{Case Study}
To intuitively demonstrate and compare the MEU capabilities of existing powerful MLLMs and our proposed method, we conducted a case study, which requires an analysis of facial dynamics between the onset and apex frames to identify the corresponding ME. As illustrated in \autoref{fig:figure8}, leading MLLMs such as GPT-4.1, Qwen-VL-MAX, and Gemini-2.5-pro exhibited difficulties in capturing the subtle dynamic variations between the two frames.
Specifically, both GPT-4.1 and Qwen-VL-MAX explicitly stated their inability to perceive the facial dynamics, and consequently, did not detect any expression. Gemini-2.5-pro, on the other hand, analyzed only the static expression and misinterpreted the emotion as "suppressed anger" or "frustration." This static-analysis approach inherently overlooks the crucial dynamic cues present across the pair of images. In this example, the frames clearly show the activation of AU12 (Lip Corner Puller), which corresponds to happiness. These observations suggest that analyzing such subtle, dynamic facial cues remains a challenging task for current MLLMs.

In contrast, our MELLM correctly identifies and explains the expression through a stepwise process. First, it performs a fine-grained motion analysis across facial ROIs to infer the activation of AU12. Second, by combining this evidence with the overall facial motion pattern, it reasons that the movement corresponds to a spontaneous smile, indicating genuine happiness. Finally, MELLM provides an explicit summary of its findings: AU12 (Lip Corner Puller) and Positive (specifically, happiness). This case study effectively demonstrates MELLM’s advanced capability to both perceive subtle dynamic cues and produce a clear, stepwise reasoning trace for its emotional inference.

\section{Conclusions}

In this paper, we present MELLM, a novel framework that bridges the gap between subtle facial motion perception and high-level semantic reasoning for Micro-Expression Understanding (MEU). Unlike traditional methods confined to discrete classification, MELLM leverages the synergy between a specialized optical flow estimator, MEFlowNet, and an LLM to interpret nuanced facial dynamics. By constructing the large-scale MEFlowDataset and the instruction-tuning MEU-Instruct dataset, we have provided the community with essential resources to facilitate research in fine-grained affective computing. Our extensive experiments demonstrate that this paradigm not only achieves state-of-the-art accuracy in cross-dataset evaluations but also offers human-readable explanations for its decisions, significantly benefiting researchers who require transparent and interpretable affective analysis.

Despite the promising capabilities demonstrated by MELLM in generalized MEU, we acknowledge certain limitations. To address these challenges, our future work will focus on three strategic directions:

\begin{enumerate}
    \item \textbf{Refining Optimization via Reinforcement Learning:} Currently, MELLM’s accuracy on the seven-class DFME in-domain test splits is marginally lower than that of the best task-specific classifiers. To bridge this gap without compromising robustness, we will explore reinforcement learning techniques. By incorporating task-specific rewards, we aim to align the LLM's reasoning process more closely with ground-truth labels.
    
    \item \textbf{Establishing Holistic Evaluation Frameworks:} Current evaluation frameworks rely primarily on traditional metrics inherited from MER.  Consequently, we plan to establish new benchmarks designed to jointly assess both the validity of the reasoning process and the correctness of the final results. 
    
    \item \textbf{Synthesizing Data for End-to-End Training:} To overcome the inherent scarcity and high acquisition costs of ME data, we will leverage parametric face models and photorealistic rendering to generate large-scale synthetic datasets equipped with fine-grained AU and expression annotations. These resources will facilitate the training of end-to-end MLLMs, paving the way for a more unified and data-efficient paradigm in MEU.
\end{enumerate}

\section*{Acknowledgments}
This work was supported by the National Natural Science Found of China (Grant No. 61727809, 62406264).

\bibliographystyle{elsarticle-num}
\bibliography{mellm} 
\end{document}